\pdfoutput=1

\documentclass{article}

\usepackage{microtype}
\usepackage{graphicx}
\usepackage{subfigure}
\usepackage{booktabs} 

\usepackage[utf8]{inputenc} 
\usepackage[T1]{fontenc}    

\usepackage{hyperref}
\hypersetup{colorlinks,linkcolor=red}

\usepackage{url}            
\usepackage{amsfonts}       
\usepackage{nicefrac}       
\usepackage{microtype}      
\usepackage{todonotes}
\usepackage{amsmath}
\usepackage{boldline}
\usepackage{wrapfig}
\usepackage{mathtools}
\usepackage{threeparttable}
\usepackage{chngcntr}

\usepackage{multibib}

\usepackage{subfiles}

\newcommand\mypara[1]{\vspace{1mm}\noindent\textbf{#1}}

\usepackage{hyperref}

\usepackage[accepted]{icml2020}

\DeclarePairedDelimiter{\ceil}{\lceil}{\rceil}

\newcommand{\Skip}[1]{{}}
\newcommand{\tocite}[1]{{}}

\newcommand{\myfig}[1]{Figure~\ref{#1}}

\newcommand{\mysec}[1]{Section~\ref{#1}}

\icmltitlerunning{Sample Factory: Egocentric 3D Control from Pixels at 100000 FPS with Asynchronous Reinforcement Learning}

\newcites{appendix}{References}

\begin{document}

\twocolumn[
\icmltitle{Sample Factory: Egocentric 3D Control from Pixels at 100000 FPS with Asynchronous Reinforcement Learning}


\begin{icmlauthorlist}
\icmlauthor{Aleksei Petrenko}{Intel,USC}
\icmlauthor{Zhehui Huang}{USC}
\icmlauthor{Tushar Kumar}{USC}
\icmlauthor{Gaurav Sukhatme}{USC}
\icmlauthor{Vladlen Koltun}{Intel}
\end{icmlauthorlist}

\icmlaffiliation{USC}{University of Southern California}
\icmlaffiliation{Intel}{Intel Labs}

\icmlcorrespondingauthor{Aleksei Petrenko}{petrenko@usc.edu}

\icmlkeywords{Machine Learning, ICML}

\icmlkeywords{Deep Reinforcement Learning, Systems, Asynchronous Reinforcement Learning, ICML}

\vskip 0.3in
]

\printAffiliationsAndNotice{}  

\begin{abstract}

Increasing the scale of reinforcement learning experiments has allowed researchers to achieve unprecedented results in both training sophisticated agents for video games, and in sim-to-real transfer for robotics. Typically such experiments rely on large distributed systems and require expensive hardware setups, limiting wider access to this exciting area of research. In this work we aim to solve this problem by optimizing the efficiency and resource utilization of reinforcement learning algorithms instead of relying on distributed computation. We present the ``Sample Factory'', a high-throughput training system optimized for a single-machine setting. Our architecture combines a highly efficient, asynchronous, GPU-based sampler with off-policy correction techniques, allowing us to achieve throughput higher than $10^5$ environment frames/second on non-trivial control problems in 3D without sacrificing sample efficiency. We extend Sample Factory to support self-play and population-based training and apply these techniques to train highly capable agents for a multiplayer first-person shooter game. Github: {\small\url{https://github.com/alex-petrenko/sample-factory}}

\end{abstract}

\section{Introduction}
\label{intro}

Training agents in simulated environments is a cornerstone of contemporary reinforcement learning research. Substantial progress has been made in recent years by applying reinforcement learning methods to train agents in these fast and efficient environments, whether it is to solve complex computer games \cite{dfp,dmquakescience, dmstarcraft2} or sophisticated robotic control problems via sim-to-real transfer \cite{rccar_sim2real, hwangbo2019learning, amdrone, dexmanip}.

Despite major improvements in the sample efficiency of modern learning methods, most of them remain notoriously data-hungry. For the most part, the level of results in recent years has risen due to the increased scale of experiments, rather than the efficiency of learning. Billion-scale experiments with complex environments are now relatively commonplace \cite{apex, impala, r2d2}, and the most advanced efforts consume trillions of environment transitions in a single training session \cite{openai2019dota}.

To minimize the turnaround time of these large-scale experiments, the common approach is to use distributed super-computing systems consisting of hundreds of individual machines \cite{openai2019dota}. Here, we show that by optimizing the architecture and improving the resource utilization of reinforcement learning algorithms, we can train agents on billions of environment transitions even on a \textit{single} compute node. We present the \textbf{``Sample Factory''}, a high-throughput training system optimized for a single-machine scenario. Sample Factory, built around an Asynchronous Proximal Policy Optimization (APPO) algorithm, is a reinforcement learning architecture that allows us to aggressively parallelize the experience collection and achieve throughput as high as 130000 FPS (environment frames per second) on a single multi-core compute node with only one GPU. We describe theoretical and practical optimizations that allow us to achieve extreme frame rates on widely available commodity hardware.

We evaluate our algorithm on a set of challenging 3D environments and demonstrate how to leverage vast amounts of simulated experience to train agents that reach high levels of skill. We then extend Sample Factory to support self-play and population-based training and apply these techniques to train highly capable agents for a full multiplayer game of Doom \cite{vizdoom}.

\section{Prior Work}
\label{prior_work}

The quest for performance and scalability has been ongoing since before the advent of deep RL \cite{mapreducerl}. Higher throughput algorithms allow for faster iteration and wider hyperparameter sweeps for the same amount of compute resources, and are therefore highly desirable.

The standard implementation of a policy gradient algorithm is fairly simple. It involves a (possibly vectorized) sampler that collects environment transitions from $N_{envs}\ge1$ copies of the environment for a fixed number of timesteps $T$. The collected batch of experience~-- consisting of $N_{envs} \times T$ samples~-- is aggregated and an iteration of SGD is performed, after which the experience can be collected again with an updated policy. This method has acquired the name Advantage Actor-Critic (A2C) in the literature \cite{doom_supercomputer}. While it is straightforward to implement and can be accelerated with batched action generation on the GPU, it has significant disadvantages. The sampling process has to halt when the actions for the next step are being calculated, and during the backpropagation step. This leads to a significant under-utilization of system resources during training. Other algorithms such as TRPO \cite{trpo} and PPO \cite{ppo} are usually also implemented in this synchronous A2C style \cite{baselines}.

Addressing the shortcomings of the naive implementation, the Asynchronous Advantage Actor-Critic (A3C) \cite{a3c} proposed a distributed scheme consisting of a number of independent actors, each with its own copy of the policy. Every actor is responsible for environment simulation, action generation, and gradient calculation. The gradients are asynchronously aggregated on a single parameter server, and actors query the updated copy of the model after each collected trajectory.

GA3C \cite{ga3c} recognized the potential of using a GPU in an asynchronous implementation for both action generation and learning. A separate learner component is introduced, and trajectories of experience are communicated between the actors and the learner instead of parameter vectors. GA3C outperforms CPU-only A3C by a significant margin, although the high communication cost between CPU actors and GPU predictors prevents the algorithm from reaching optimal performance. 

IMPALA \cite{impala} uses an architecture conceptually similar to GA3C, extended to support distributed training. An efficient implementation of GPU batching for action generation leads to increased throughput, with reported training frame rate of 24K FPS for a single machine with 48 CPU cores, and up to 250K FPS on a cluster with 500 CPUs.

The need for ever larger-scale experiments has focused attention on  high-throughput reinforcement learning in recent publications. Decentralized Distributed PPO \cite{ddppo} optimizes the distributed policy gradient setup for multi-GPU clusters and resource-intensive environments by parallelizing the learners and significantly reducing the network throughput required.
Concurrent with this work, SEED RL \cite{seedrl} improves upon the IMPALA architecture and achieves high throughput in both single-machine and multi-node scenarios, although unlike Sample Factory it focuses on more expensive hardware setups involving multiple accelerators.

Deep RL frameworks also provide high-throughput implementations of policy gradient algorithms. RLlib \cite{rllib}, based on the distributed computation framework Ray \cite{ray}, and TorchBeast \cite{torchbeast} provide optimized implementations of the IMPALA architecture. Rlpyt \cite{rlpyt} implements highly-efficient asynchronous GPU samplers that share some ideas with our work, although currently it does not include asynchronous policy gradient methods such as IMPALA or APPO.

Methods such as APE-X \cite{apex} and R2D2 \cite{r2d2} demonstrate the great scalability of off-policy RL. While off-policy algorithms exhibit state-of-the-art performance in domains such as Atari \cite{ale}, they may be difficult to extend to the full complexity of more challenging problems \cite{dmstarcraft2}, since Q-functions may be hard to learn for large multi-headed and autoregressive action spaces. In this work, we focused on policy gradient methods, although there is great potential in off-policy learning. Hybrid methods such as LASER \cite{laser} promise to combine high scalability, flexibility, and sample efficiency.

\begin{figure*}[t]
  \centering
  {\small
  \def\svgwidth{0.68\textwidth}
  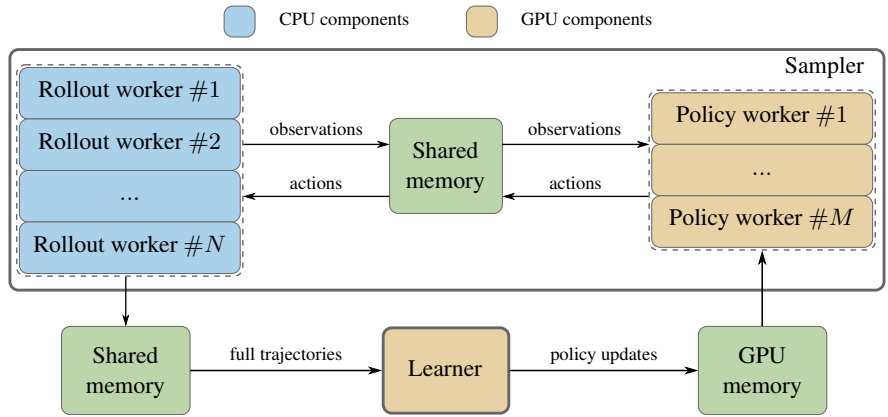
  \vspace{-5pt}
   \caption{Overview of the Sample Factory architecture. $N$ parallel rollout workers simulate $k$ environments each, collecting observations. These observations are processed by $M$ policy workers, which generate actions and new hidden states via an accelerated forward pass on the GPU. Complete trajectories are sent from rollout workers to the learner. After the learner completes the backpropagation step, the model parameters are updated in shared CUDA memory and immediately fetched by the policy workers.}
   \vspace{-5pt}
    \label{fig:algorithm_design}
  }
\end{figure*}
\section{Sample Factory}
\label{method}

Sample Factory is an architecture for high-throughput reinforcement learning on a single machine. When designing the system we focused on making all key computations fully asynchronous, as well as minimizing the latency and the cost of communication between components, taking full advantage of fast local messaging.

A typical reinforcement learning scenario involves three major computational workloads: environment simulation, model inference, and backpropagation. Our key motivation was to build a system in which the slowest of three workloads never has to wait for any other processes to provide the data necessary to perform the next computation, since the overall throughput of the algorithm is ultimately defined by the workload with the \textit{lowest} throughput. In order to minimize the amount of time processes spend waiting, we need to guarantee that the new portion of the input is always available, even before the next step of computation is about to start. The system in which the most compute-intensive workload never idles can reach the highest resource utilization, thereby approaching optimal performance.

\subsection{High-level design}

The desire to minimize the idle time for all key computations motivates the high-level design of the system (\myfig{fig:algorithm_design}). We associate each computational workload with one of three dedicated types of components. These components communicate with each other using a fast protocol based on FIFO queues and shared memory. The queueing mechanism provides the basis for continuous and asynchronous execution, where the next computation step can be started immediately as long as there is something in the queue to process. The decision to assign each workload to a dedicated component type also allows us to parallelize them independently, thereby achieving optimized resource balance. This is different from prior work \cite{a3c, impala}, where a single system component, such as an actor, typically has multiple responsibilities. The three types of components involved are rollout workers, policy workers, and learners.

\textbf{\textit{Rollout workers}} are solely responsible for environment simulation. Each rollout worker hosts $k \ge 1$ environment instances and sequentially interacts with these environments, collecting observations $x_t$ and rewards $r_t$. Note that the rollout workers do not have their own copy of the policy, which makes them very lightweight, allowing us to massively parallelize the experience collection on modern multi-core CPUs.

The observations $x_t$ and the hidden states of the agent $h_t$ are then sent to the \textbf{\textit{policy worker}}, which collects batches of $x_t, h_t$ from multiple rollout workers and calls the policy $\pi$, parameterized by the neural network $\theta_\pi$ to compute the action distributions $\mu(a_t|x_t, h_t)$, and the updated hidden states $h_{t+1}$. The actions $a_t$ are then sampled from the distributions $\mu$, and along with $h_{t+1}$ are communicated back to the corresponding rollout worker. This rollout worker uses the actions $a_t$ to advance the simulation and collect the next set of observations $x_{t+1}$ and rewards $r_{t+1}$.

Rollout workers save every environment transition to a trajectory buffer in shared memory. Once $T$ environment steps are simulated, the trajectory of observations, hidden states, actions, and rewards $\tau = x_1, h_1, a_1, r_1, ..., x_T, h_T, a_T, r_T$ becomes available to the \textbf{\textit{learner}}. The learner continuously processes batches of trajectories and updates the parameters of the actor $\theta_\pi$ and the critic $\theta_V$. These parameter updates are sent to the policy worker as soon as they are available, which reduces the amount of experience collected by the previous version of the model, minimizing the average policy lag. This completes one training iteration.

\mypara{Parallelism.} As mentioned previously, the rollout workers do not own a copy of the policy and therefore are essentially thin wrappers around the environment instances. This allows them to be massively parallelized. Additionally, Sample Factory also parallelizes policy workers. This can be achieved because all of the current trajectory data ($x_t, h_t, a_t, ...$) is stored in shared tensors that are accessible by all processes. This allows the policy workers themselves to be stateless, and therefore consecutive trajectory steps from a single environment can be easily processed by any of them. In practical scenarios, $2$ to $4$ policy worker instances easily saturate the rollout workers with actions, and together with a special sampler design (section \ref{sec:sampling}) allow us to eliminate this potential bottleneck.

The learner is the only component of which we run a single copy, at least as long as single-policy training is concerned (multi-policy training is discussed in section \ref{sec:multi-agent}). We can, however, utilize multiple accelerators on the learner through data-parallel training and Hogwild-style parameter updates \cite{hogwild}. Together with large batch sizes typically required for stable training in complex environments, this gives the learner sufficient throughput to match the experience collection rate, unless the computational graph is highly non-trivial.

\subsection{Sampling}
\label{sec:sampling}

Rollout workers and policy workers together form the sampler. The sampling subsystem most critically affects the throughput of the RL algorithm, since it is often the bottleneck. We propose a specific way of implementing the sampler that allows for optimal resource utilization through minimizing the idle time on the rollout workers.

First, note that training and experience collection are decoupled, so new environment transitions can be collected during the backpropagation step. There are no parameter updates for the rollout workers either, since the job of action generation is off-loaded to the policy worker. However, if not addressed, this still leaves the rollout workers waiting for the actions to be generated by policy workers and transferred back through interprocess communication.

\begin{figure}
  \centering
  {\tiny
  \def\svgwidth{1.0\linewidth}
  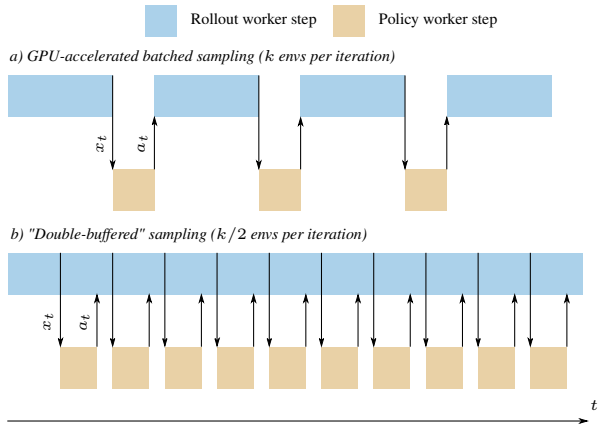
  }
  \vspace{-15pt}
  \caption{a) Batched sampling enables forward pass acceleration on GPU, but rollout workers have to wait for actions before the next environment step can be simulated, underutilizing the CPU. b) Double-buffered sampling splits $k$ environments on the rollout worker into two groups, alternating between them during sampling, which practically eliminates idle time on CPU workers.
  }
  \vspace{-12pt}
  \label{fig:double_buffered}
\end{figure}

To alleviate this inefficiency we use Double-Buffered Sampling (Figure \ref{fig:double_buffered}). Instead of storing only a single environment on the rollout worker, we instead store a vector of environments $E_1, ..., E_k$, where $k$ is even for simplicity. We split this vector into two groups $E_1, ..., E_{k/2}$, $E_{k/2 + 1}, ..., E_{k}$, and alternate between them as we go through the rollout. While the first group of environments is being stepped through, the actions for the second group are calculated on the policy worker, and vice versa. With a fast enough policy worker and a correctly tuned value for $k$ we can completely mask the communication overhead and ensure full utilization of the CPU cores during sampling, as illustrated in \myfig{fig:double_buffered}. For maximal performance with double-buffered sampling we want $k/2 > \ceil[\big]{{t_{inf}}/{t_{env}}}$, where $t_{inf}$ and $t_{env}$ are average inference and simulation time, respectively.

\subsection{Communication between components}

The key to unlocking the full potential of the local, single-machine setup is to utilize fast communication mechanisms between system components. As suggested by \myfig{fig:algorithm_design}, there are four main pathways for information flow: two-way communication between rollout and policy workers, transfer of complete trajectories to the learner, and transfer of parameter updates from the learner to the policy worker. For the first three interactions we use a mechanism based on PyTorch \cite{pytorch} shared memory tensors. We note that most data structures used in an RL algorithm can be represented as tensors of fixed shape, whether they are trajectories, observations, or hidden states. Thus we preallocate a sufficient number of tensors in system RAM. Whenever a component needs to communicate, we copy the data into the shared tensors, and send only the indices of these tensors through FIFO queues, making messages tiny compared to the overall amount of data transferred.

For the parameter updates we use memory sharing on the GPU. Whenever a model update is required, the policy worker simply copies the weights from the shared memory to its local copy of the model.

Unlike many popular asynchronous and distributed implementations, we do not perform any kind of data serialization as a part of the communication protocol. At full throttle, Sample Factory generates and consumes more than 1 GB of data per second, and even the fastest serialization/deserialization mechanism would severely hinder throughput.

\subsection{Policy lag}

Policy lag is an inherent property of asynchronous RL algorithms, a discrepancy between the policy that collected the experience (\textit{behavior policy}) and the \textit{target} policy that is learned from it. The existence of this discrepancy conditions the off-policy training regime. Off-policy learning is known to be hard for policy gradient methods, in which the model parameters are usually updated in the direction of $\nabla \log \mu (a_s \vert x_s) q(x_s, a_s)$, where $q(x_s, a_s)$ is an estimate of the policy state-action value. The bigger the policy lag, the harder it is to correctly estimate this gradient using a set of samples $x_s$ from the behavior policy. Empirically this gets more difficult in learning problems that involve recurrent policies, high-dimensional observations, and complex action spaces, in which even very similar policies are unlikely to exhibit the same performance over a long trajectory.

Policy lag in an asynchronous RL method can be caused either by acting in the environment using an old policy, or collecting more trajectories from parallel environments in one iteration than the learner can ingest in a single minibatch, resulting in a portion of the experience becoming off-policy by the time it is processed. We deal with the first issue by immediately updating the model on policy workers, as soon as new parameters become available. In Sample Factory the parameter updates are cheap because the model is stored in shared memory. A typical update takes less than 1 ms, therefore we collect a very minimal amount of experience with a policy that is different from the ``master'' copy.

It is however not necessarily possible to eliminate the second cause. It is beneficial in RL to collect training data from many environment instances in parallel. Not only does this decorrelate the experiences, it also allows us to utilize multi-core CPUs, and with larger values for $k$ (environments per core), take full advantage of the double-buffered sampler. In one ``iteration'' of experience collection, $n$ rollout workers, each running $k$ environments, will produce a total of $N_{iter}=n \times k \times T$ samples. Since we update the policy workers immediately after the learner step, potentially in the middle of a trajectory, this leads to the earliest samples in trajectories lagging behind $N_{iter} / N_{batch} - 1$ policy updates on average, while the newest samples have no lag.

One can minimize the policy lag by decreasing $T$ or increasing the minibatch size $N_{batch}$. Both have implications for learning. We generally want larger $T$, in the $2^5$--$2^7$ range for backpropagation through time with recurrent policies, and large minibatches may reduce sample efficiency. The optimal batch size depends on the particular environment, and larger batches were shown to be suitable for complex problems with noisy gradients \cite{large_batch_openai_sam_mccandlish2018}.

Additionally, there are two major classes of techniques designed to cope with off-policy learning. The first idea is to apply trust region methods \cite{trpo, ppo}: by staying close to the behavior policy during learning, we improve the quality of gradient estimates obtained using samples from this policy. Another approach is to use importance sampling to correct the targets for the value function $V^\pi$ to improve the approximation of the discounted sum of rewards under the \textit{target} policy \cite{retrace}. IMPALA \cite{impala} introduced the V-trace algorithm that uses truncated importance sampling weights to correct the value targets. This was shown to improve the stability and sample-efficiency of off-policy learning.

Both methods can be applied independently, as V-trace corrects our training objective and the trust region guards against destructive parameter updates. Thus we implemented both V-trace and PPO clipping in Sample Factory. Whether to use these methods or not can be considered a hyperparameter choice for a specific experiment. We find that a combination of PPO clipping and V-trace works well across tasks and yields stable training, therefore we decided to use both methods in all experiments reported in the paper.

\begin{figure*}[ht]
\vspace{-5pt}
\centering
\includegraphics[scale=0.6]{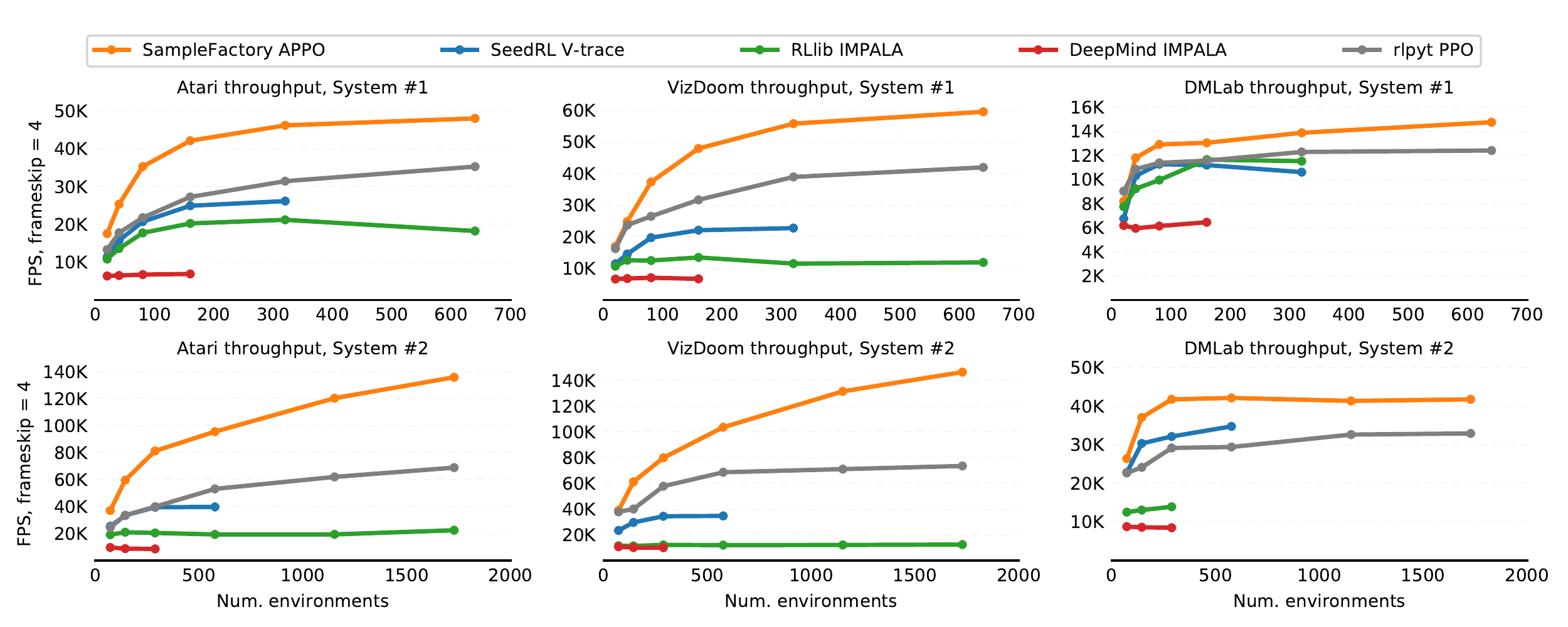}
\vspace{-5pt}
\caption{Training throughput, measured in environment frames per second.}
\vspace{-5pt}
\label{fig:throughput}
\end{figure*}

\subsection{Multi-agent learning and self-play}
\label{sec:multi-agent}

Some of the most advanced recent results in deep RL have been achieved through multi-agent reinforcement learning and self-play \cite{self-play, openai2019dota}. Agents trained via self-play are known to exhibit higher levels of skill than their counterparts trained in fixed scenarios \cite{dmquakescience}. As policies improve during self-play they generate a training environment of gradually increasing complexity, naturally providing a curriculum for the agents and allowing them to learn progressively more sophisticated skills. Complex behaviors (e.g.\ cooperation and tool use) have been shown to emerge in these training scenarios \cite{hide-n-seek}.

There is also evidence that \textit{populations} of agents training together in multi-agent environments can avoid some failure modes experienced by regular self-play setups, such as early convergence to local optima or overfitting. A diverse training population can expose agents to a wider set of adversarial policies and produce more robust agents, reaching higher levels of skill in complex tasks \cite{dmstarcraft2, dmquakescience}.

To unlock the full potential of our system we add support for multi-agent environments, as well as training populations of agents. Sample Factory naturally extends to multi-agent and multi-policy learning. Since the rollout workers are mere wrappers around the environment instances, they are totally agnostic to the policies providing the actions. Therefore to add more policies to the training process we simply spawn more policy workers and more learners to support them. On the rollout workers, for every agent in every multi-agent environment we sample a random policy $\pi_i$ from the population at the beginning of each episode. The action requests are then routed to their corresponding policy workers using a set of FIFO queues, one for every $\pi_i$. The population-based setup that we use in this work is explained in more detail in \mysec{experiments}.

\section{Experiments}
\label{experiments}

\subsection{Computational performance}

Since increasing throughput and reducing experiment turnaround time was the major motivation behind our work, we start by investigating the computational aspects of system performance. We measure training frame rate on two hardware systems that closely resemble commonly available hardware setups in deep learning research labs. In our experiments, System $\#1$ is a workstation-level PC with a 10-core CPU and a GTX 1080 Ti GPU. System $\#2$ is equipped with a server-class 36-core CPU and a single RTX 2080 Ti.

As our testing environments we use three simulators: Atari \cite{ale}, VizDoom \cite{vizdoom}, and DeepMind Lab \cite{dmlab}. While that Atari Learning Environment is a collection of 2D pixel-based arcade games, VizDoom and DMLab are based on the rendering engines of immersive 3D first-person games, Doom and Quake III. Both VizDoom and DMLab feature first-person perspective, high-dimensional pixel observations, and rich configurable training scenarios. For our throughput measurements in Atari we used the game Breakout, with grayscale frames in $84 \times 84$ resolution and 4-framestack. In VizDoom we chose the environment \textit{Battle} described in section \ref{sec:vizdoom}, with the observation resolution of $128 \times 72 \times 3$. Finally, for DeepMind Lab we used the environment \textit{rooms\_collect\_good\_objects} from DMLab-30, also referred to as \textit{seekavoid\_arena\_01} \cite{impala}. The resolution for DeepMind Lab is kept at standard $96 \times 72 \times 3$. We follow the original implementation of IMPALA and use a CPU-based software renderer for Lab environments. We noticed that higher frame rate can be achieved when using GPUs for environment rendering, especially on System $\#1$ (see appendix). The reported throughput is measured in simulated environment steps per second, and in all three testing scenarios we used traditional 4-frameskip, where the RL algorithm receives a training sample every 4 environment steps.

We compare performance of Sample Factory to other high-throughput policy gradient methods. Our first baseline is an original version of the IMPALA algorithm \cite{impala}. The second baseline is IMPALA implemented in RLlib \cite{rllib}, a high-performance distributed RL framework. Third is a recent evolution of IMPALA from DeepMind, SeedRL \cite{seedrl}. Our final comparison is against a version of PPO with asynchronous sampling from the rlpyt framework \cite{rlpyt}, one of the fastest open-source RL implementations. We use the same model architecture for all methods, a ConvNet with three convolutional layers, an RNN core, and two fully-connected heads for the actor and the critic. Full benchmarking details, including hardware configuration and model architecture are provided in the supplementary files.

Figure \ref{fig:throughput} illustrates the training throughput in different configurations averaged over five minutes of continuous training to account for performance fluctuations caused by episode resets and other factors. Aside from showing the peak frame rate we also demonstrate how the performance scales with the increased number of environments sampled in parallel.

Sample Factory outperforms the baseline methods in most of the training scenarios. Rlpyt and SeedRL follow closely, matching Sample Factory performance in some configurations with a small number of environments. Both IMPALA implementations fail to efficiently utilize the resources in a single-machine deployment and hit performance bottlenecks related to data serialization and transfer. Additionally, their higher per-actor memory usage did not allow us to sample as many environments in parallel. We omitted data points for configurations that failed due to lack of memory or other resources.

\begin{figure}[h]
\centering
\includegraphics[width=\linewidth]{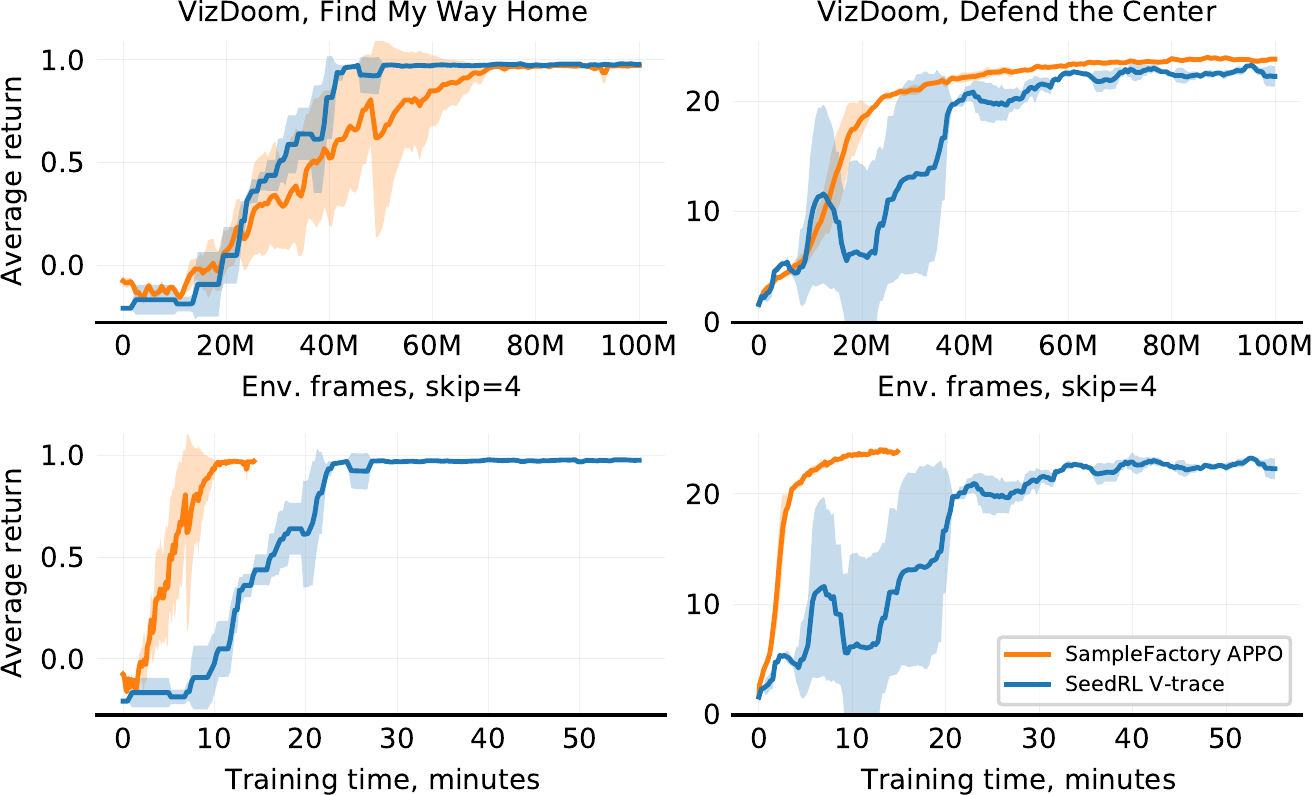}
\vspace{-10pt}
\caption{Direct comparison of wall-time performance. We show the mean and standard deviation of four training runs for each experiment.}
\vspace{-15pt}
\label{fig:wall-time}
\end{figure}

Figure \ref{fig:wall-time} demonstrates how the system throughput translates into raw wall-time training performance. Sample Factory and SeedRL implement similar asynchronous architectures and demonstrate very close sample efficiency with equivalent sets of hyperparameters. We are therefore able to compare the training time directly. We trained agents on two standard VizDoom environments. The plots demonstrate a 4x advantage of Sample Factory over the state-of-the-art baseline. Note that direct fair comparison with the fastest baseline, rlpyt, is not possible since it does not implement asynchronous training. In rlpyt the learner waits for all workers to finish their rollouts before each iteration of SGD, therefore increasing the number of sampled environments also increases the training batch size, which significantly affects sample efficiency. This is not the case for SeedRL and Sample Factory, where a fixed batch size can be used regardless of the number of environments simulated.

\begin{table}[b]
\centering
\setlength{\tabcolsep}{3mm}
\vspace{-10pt}
\tiny
\begin{tabular}{l c c c}

    & Atari, FPS & VizDoom, FPS & DMLab, FPS \\
    \midrule
    
    Pure simulation & 181740 (100\%) & 322907 (100\%) & 49679 (100\%) \\ 
    \midrule
    DeepMind IMPALA & 9961 (5.3\%) & 10708 (3.3\%) & 8782 (17.7\%) \\
    RLlib IMPALA & 22440 (12.3\%) & 12391 (3.8\%) & 13932 (28.0\%) \\ 
    SeedRL V-trace & 39726 (21.9\%) & 34428 (10.7\%) & 34773 (70.0\%) \\ 
    rlpyt PPO & 68880 (37.9\%) & 73544 (22.8\%) & 32948 (66.3\%) \\ 
    \midrule
    SampleFactory APPO & \textbf{135893 (74.8\%)} & \textbf{146551 (45.4\%)} & \textbf{42149 (84.8\%)} \\ 
\end{tabular}
\caption{Peak throughput of various RL algorithms on System $\#2$ in environment frames per second and as percentage of the optimal frame rate.}
\label{tab:rooflining}
\end{table}

Finally, we also analyzed the theoretical limits of RL training throughput. By stripping away all computationally expensive workloads from our system we can benchmark a bare-bones sampler that just executes a random policy in the environment as quickly as possible. The framerate of this sampler gives us an upper bound on training performance, emulating an ideal RL algorithm with infinitely fast action generation and learning. Table \ref{tab:rooflining} shows that Sample Factory gets significantly closer to this ideal performance than the baselines. This experiment also shows that further optimization may be possible. For VizDoom, for example, the sampling rate is so high that the learner loop completely saturates the GPU even with relatively shallow models. Therefore performance can be further improved by using multiple GPUs in data-parallel mode, or, alternatively, we can train small populations of agents, with learner processes of different policies spread across GPUs.

\subsection{DMLab-30 experiment}

\begin{figure}[h]
\centering
\includegraphics[width=0.8\linewidth]{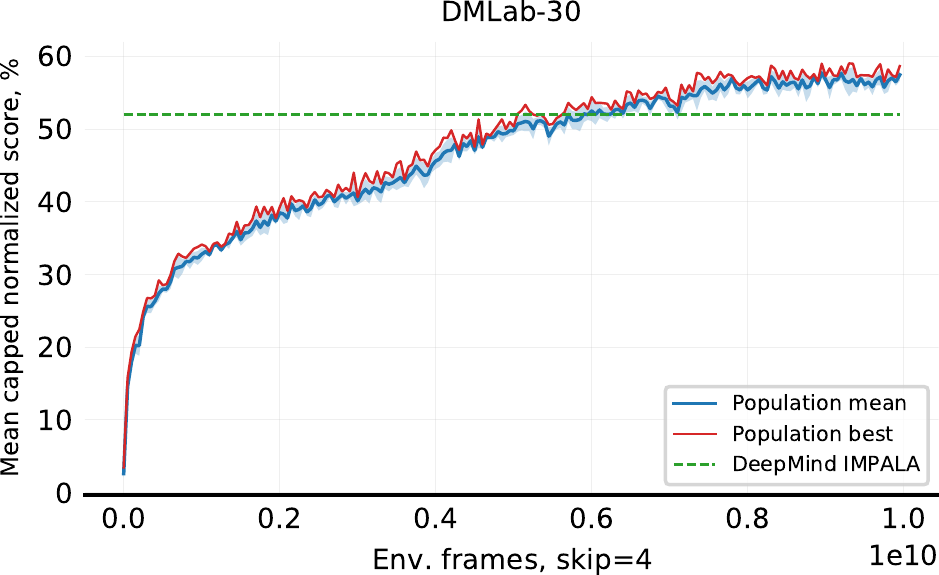}
\vspace{-5pt}
\caption{Mean capped human-normalized training score \cite{impala} for a \emph{single-machine} DMLab-30 PBT run with Sample Factory. (Compared to cluster-scale IMPALA deployment.)}
\vspace{0pt}
\label{fig:dmlab-30}
\end{figure}

IMPALA \cite{impala} showed that with sufficient computational power it is possible to move beyond single-task RL and train one agent to solve a set of 30 diverse pixel-based environments at once. Large-scale multi-task training can facilitate the emergence of complex behaviors, which motivates further investment in this research direction. To demonstrate the efficiency and flexibility of Sample Factory we use our system to train a population of four agents on DMLab-30 (Figure \ref{fig:dmlab-30}). While the original implementation relied on a distributed multi-server setup, our agents were trained on a single 36-core 4-GPU machine.
Sample Factory reduces the computational requirements for large-scale experiments and makes multi-task benchmarks like DMLab-30 accessible to a wider research community. To support future research, we also release a dataset of pre-generated environment layouts for DMLab-30 which contains a sufficient number of unique environments for $10^{10}$-sample training and beyond. This dataset removes the need to dynamically generate new layouts during training, which leads to a multifold increase in throughput on DMLab-30.

\subsection{VizDoom experiments}
\label{sec:vizdoom}

\begin{figure*}[h]
\centering
\includegraphics[width=\textwidth]{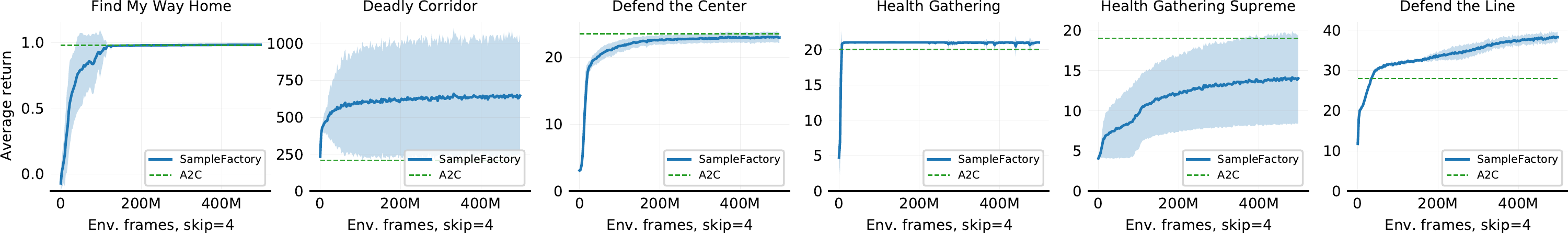}
\vspace{-10pt}
\caption{Training curves for standard VizDoom scenarios. We show the mean and standard deviation for ten independent experiments conducted for each scenario.}
\vspace{-5pt}
\label{fig:basic_envs}
\end{figure*}

We further use Sample Factory to train agents on a set of VizDoom environments. VizDoom provides challenging scenarios with very high potential skill cap. It supports rapid experience collection at fairly high input resolution. With Sample Factory, we can train agents on \emph{billions} of environment transitions in a matter of hours (see \myfig{fig:throughput}). Despite substantial effort put into improving VizDoom agents, including several years of AI competitions, the best reported agents are still far from reaching expert-level human performance \cite{doom_competitions}.

We start by examining agent performance in a set of basic environments included in the VizDoom distribution (Figure \ref{fig:basic_envs}). Our algorithm matches or exceeds the performance reported in prior work on the majority of these tasks \cite{doom_supercomputer}.

\begin{figure}[b!]
\centering
\includegraphics[width=\linewidth]{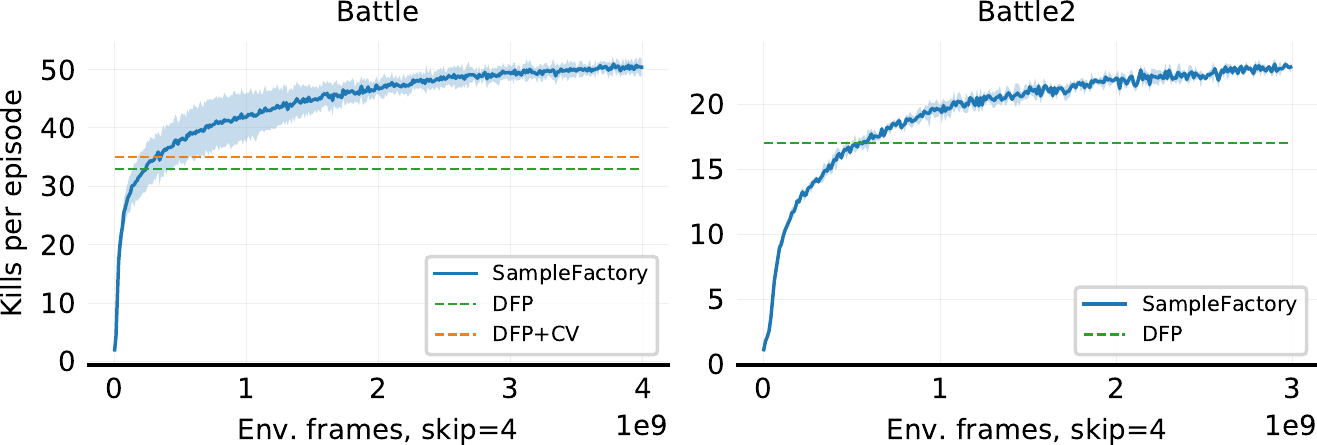}
\vspace{-5pt}
\caption{VizDoom battle experiments. We show the mean and standard deviation for four independent runs. Here as baselines we provide scores reported for Direct Future Prediction (DFP) \cite{dfp}, and a version of DFP with additional input modalities such as depth and segmentation masks, produced by a computer vision subsystem \cite{cv_matter_for_action}. The latter work only reports results for \textit{Battle}.}
\vspace{0pt}
\label{fig:single-player-envs}
\end{figure}

We then investigate the performance of Sample Factory agents in four advanced single-player game modes: \textit{Battle}, \textit{Battle2}, \textit{Duel}, and \textit{Deathmatch}. In \textit{Battle} and \textit{Battle2}, the goal of the agent is to defeat adversaries in an enclosed maze while maintaining health and ammunition. The maze in \textit{Battle2} is a lot more complex, with monsters and healthpacks harder to find. The action set in the battle scenarios includes five independent discrete action heads for moving, aiming, strafing, shooting, and sprinting. As shown in \myfig{fig:single-player-envs}, our final scores on these environments significantly exceed those reported in prior work \cite{dfp, cv_matter_for_action}.

We also introduce two new environments, \textit{Duel} and \textit{Deathmatch}, based on popular large multiplayer maps often chosen for competitive matches between human players. Single-player versions of these environments include scripted in-game opponents (bots) and can thus emulate a full Doom multiplayer gameplay while retaining high single-player simulation speed. We used in-game opponents that are included in standard Doom distributions. These bots are programmed by hand and have full access to the environment state, unlike our agents, which only receive pixel observations and auxiliary info such as the current levels of health and ammunition.

\begin{figure}[b!]
\centering
\includegraphics[width=\linewidth]{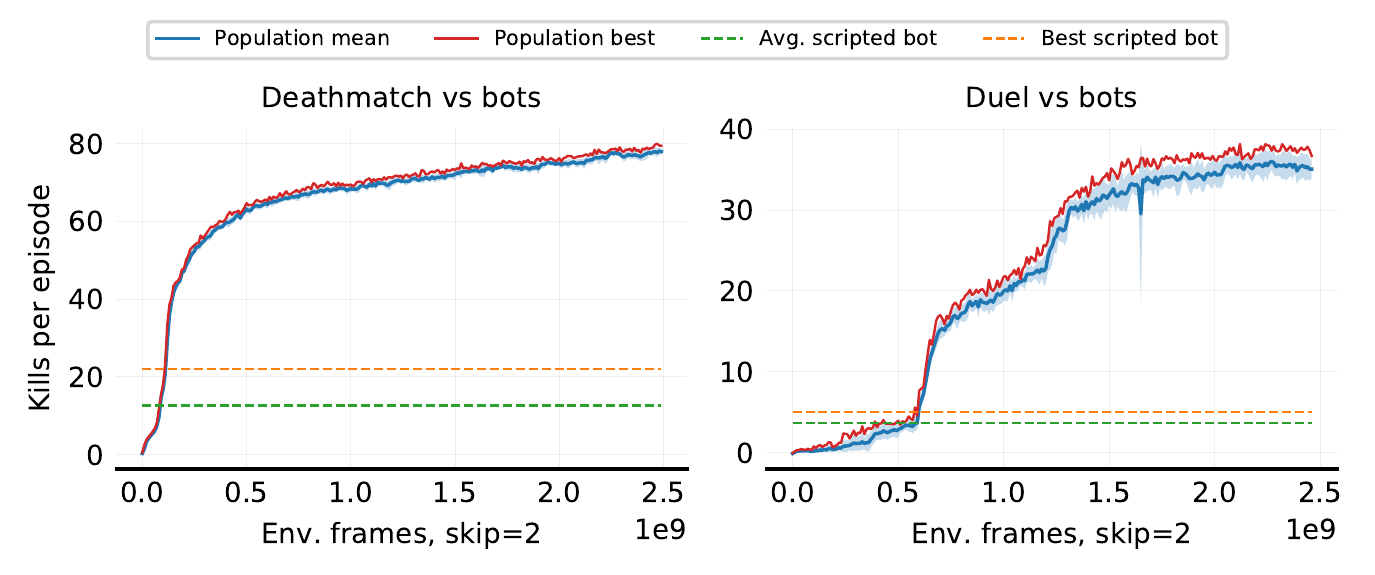}
\vspace{-5pt}
\caption{Populations of 8 agents trained in \textit{Deathmatch} and \textit{Duel} scenarios. On the y-axis we report the average number of adversaries defeated in an 4-minute match. Shown are the means and standard deviations within the population, as well as the performance of the best agent. }
\vspace{-5pt}
\label{fig:self-play}
\end{figure}

For \textit{Duel} and \textit{Deathmatch} we extend the action space to also include weapon switching and object interaction, which allows the agent to open doors and call elevators. The augmented action space fully replicates a set of controls available to a human player. This brings the total number of possible actions to $\sim 1.2 \times 10^4$, which makes the policies significantly more complex than those typically used for Atari or DMLab. We find that better results can be achieved in these environments when we repeat actions for two consecutive frames instead of the traditional four \cite{ale}, allowing the agents to develop precise movement and aim. In \textit{Duel} and \textit{Deathmatch} experiments we use a 36-core PC with four GPUs to harness the full power of Sample Factory and train a population of 8 agents with population-based training. The final agents beat the in-game bots on the highest difficulty in 100\% of the matches in both environments. In \textit{Deathmatch} our agents defeat scripted opponents with an average score of $80.5$ versus $12.6$. In \textit{Duel} the average score is $34.7$ to $3.6$ frags per episode (Figure \ref{fig:self-play}).

\mypara{Self-play experiment.} Using the networking capabilities of VizDoom we created a Gym interface \cite{gym} for full multiplayer versions of \textit{Duel} and \textit{Deathmatch} environments. In our implementation we start a separate environment instance for every participating agent, after which these environments establish network connections using UDP sockets. The simulation proceeds one step at a time, synchronizing the state between the game instances connected to the same match through local networking. This environment allows us to evaluate the ultimate configuration of Sample Factory, which includes both multi-agent and population-based training.

We use this configuration to train a population of eight agents playing against each other in 1v1 matches in a \textit{Duel} environment, using a setup similar to the ``For The Win'' (FTW) agent described in \cite{dmquakescience}. As in scenarios with scripted opponents, within one episode our agents optimize environment reward based on game score and in-game events, including positive reinforcement for scoring a kill or picking up a new weapon and penalties for dying or losing armor. The agents are meta-optimized through hyperparameter search via population-based training. The meta-objective in the self-play case is simply winning, with a reward of $+1$ for outscoring the opponent and $0$ for any other outcome. This is different from our experiments with scripted opponents, where the final objective for PBT was based on the total number of kills, because agents quickly learned to win 100\% of the matches against scripted bots.

During population-based training we randomly mutate the bottom 70\% of the population every $5\times10^6$ environment frames, altering hyperparameters such as learning rate, entropy coefficient, and reward weights. If the win rate of the policy is less than half of the best-performing agent's win rate, we simply copy the model weights and hyperparameters from the best agent to the underperforming agent and continue training.

\begin{figure}[ht]
\centering
\begin{tabular}{ c }
    \includegraphics[width=0.8\linewidth]{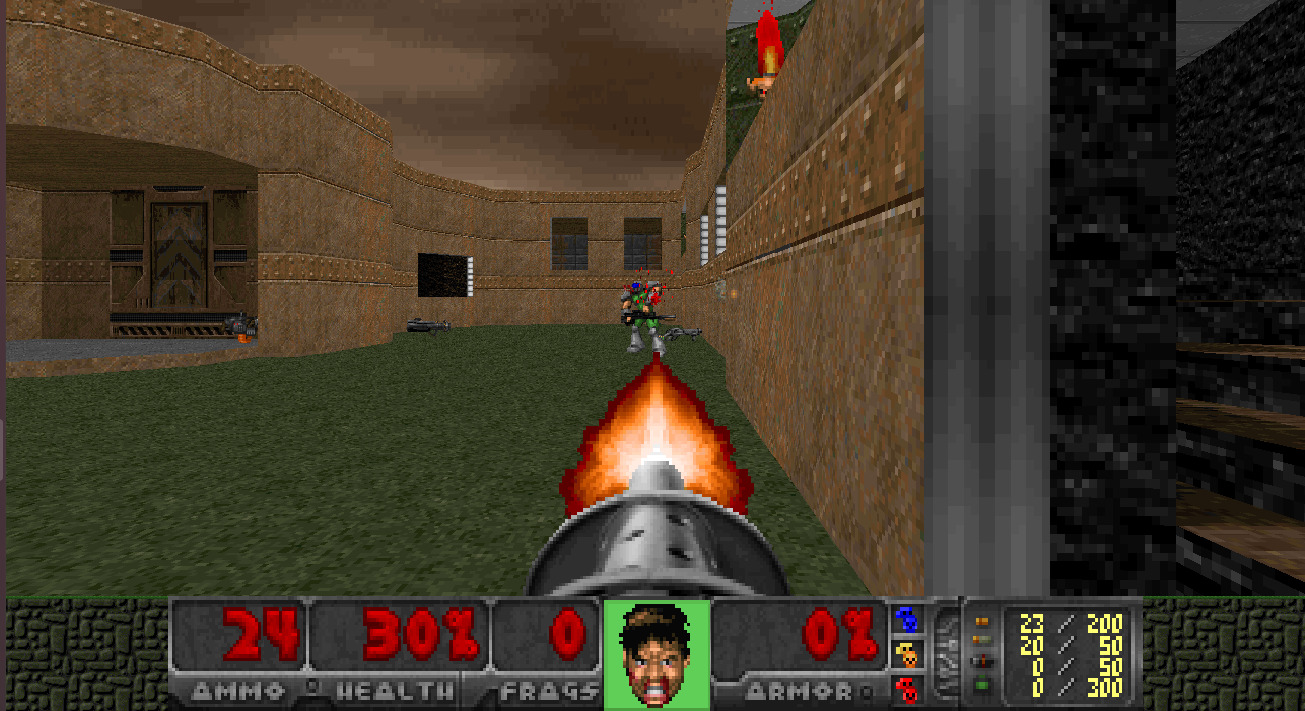} \\ \includegraphics[width=0.8\linewidth]{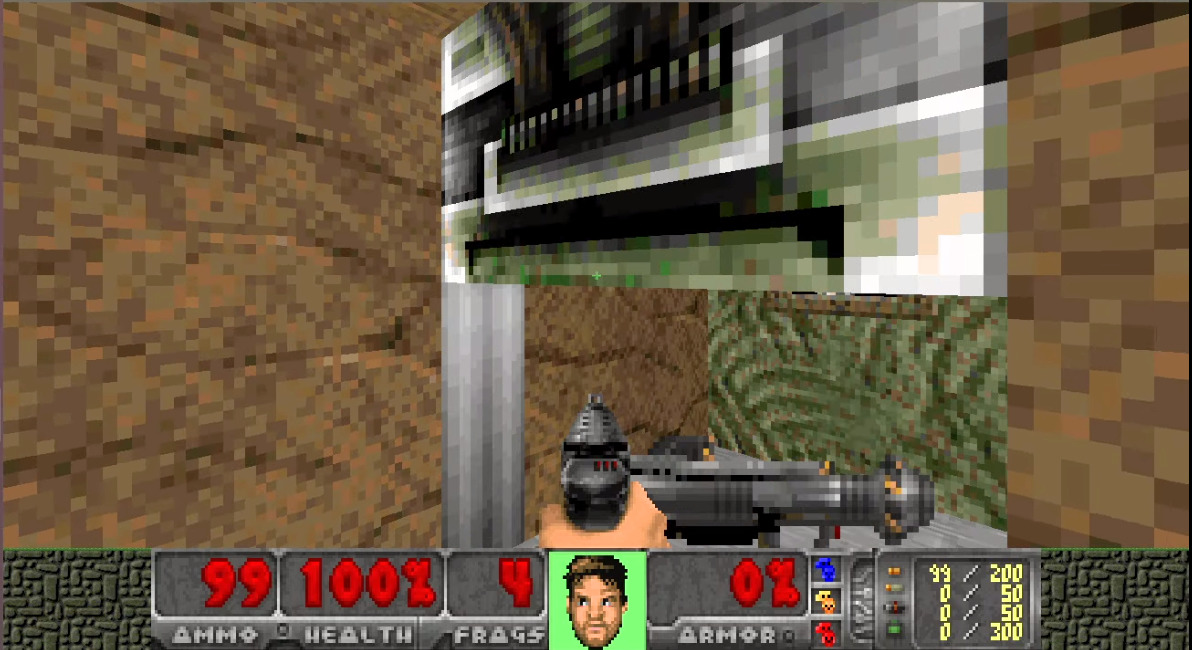} 
\end{tabular}
\vspace{-5pt}
\caption{Behavior of an agent trained via self-play. \textbf{Top:} Agents tend to choose the chaingun to shoot at their opponents from longer distance. \textbf{Bottom:} Agent opening a secret door to get a more powerful weapon.}
\label{fig:doom-self-play}
\vspace{-10pt}
\end{figure}

As in our experiments with scripted opponents, each of the eight agents was trained for $2.5 \times 10^9$ environment frames on a single 36-core 4-GPU server, with the whole population consuming $\sim18$ years of simulated experience. We observe that despite a relatively small population size, a diverse set of strategies emerges. We then simulated 100 matches between the self-play (FTW) agent and the agent trained against scripted bots, selecting the agent with the highest score from both populations. The results were 78 wins for the self-play agent, 3 losses, and 19 ties. This demonstrates that population-based training resulted in more robust policies (Figure \ref{fig:doom-self-play}), while the agent trained against bots ultimately overfitted to a single opponent type. Video recordings of our agents can be found at {\small\url{https://sites.google.com/view/sample-factory}}.

\section{Discussion}
\label{discussion}

We presented an efficient high-throughput reinforcement learning architecture that can process more than $10^5$ environment frames per second on a single machine. We aim to democratize deep RL and make it possible to train whole populations of agents on billions of environment transitions using widely available commodity hardware. We believe this is an important area of research, as it can benefit any project that leverages model-free RL. With our system architecture, researchers can iterate on their ideas faster, thus accelerating progress in the field.

We also want to point out that maximizing training efficiency on a single machine is equally important for \textit{distributed} systems. In fact, Sample Factory can be used as a single node in a distributed setup, where each machine has a sampler and a learner. The learner computes gradients based on locally collected experience only, and learners on multiple nodes can then synchronize their parameter updates after every training iteration, akin to DD-PPO \cite{ddppo}.

We showed the potential of our architecture by training highly capable agents for a multiplayer configuration of the immersive 3D game Doom. We chose the most challenging scenario that exists in first-person shooter games~-- a duel. Unlike multiplayer deathmatch, which tends to be chaotic, the duel mode requires strategic reasoning, positioning, and spatial awareness. Despite the fact that our agents were able to convincingly defeat scripted in-game bots of the highest difficulty, they are not yet at the level of expert human players. One of the advantages human players have in a duel is the ability to perceive sound. An expert human player can hear the sounds produced by the opponent (ammo pickups, shots fired, etc.) and can integrate these signals to determine the opponent's position. Recent work showed that RL agents can beat humans in pixel-based 3D games in limited scenarios \cite{dmquakescience}, but the task of defeating expert competitors in a full game, as played by humans, requires additional research, for example into fusing information from multiple sensory systems.

\bibliography{bibtex}
\bibliographystyle{icml2020}

\onecolumn
\icmltitle{Supplementary Material}

\vskip 0.3in

\appendix

\counterwithin{figure}{section}
\counterwithin{table}{section}

\section{Experimental Details}
\subsection{Performance analysis}

In this section, we provide details of the experimental setup used in our performance benchmarks. One of the goals of our experiments was to compare the performance of different asynchronous RL algorithms "apples to apples", i.e. where all the details that influence throughput are exactly the same for all methods we compare. This includes hardware configuration, simulated environments and their settings (e.g. observation resolution), model size and architecture, and the number of environment instances sampled in parallel.

\subsubsection{Hardware configuration}

We focused on commodity hardware often used for deep learning experimentation. Systems $\#1$ and $\#2$ were used for performance benchmarks. System $\#3$ is similar to System $\#2$, except with four accelerators instead of one. We used System $\#3$ for our large-scale experiments with self-play and population-based training. See Table \ref{tab:hw} for details.

\begin{table*}[h]
\centering
\setlength{\tabcolsep}{3mm}
\small
\begin{tabular}{l @{\hspace{4em}} c @{\hspace{2em}} c @{\hspace{2em}} c}
    \toprule
    & System $\#1$ & System $\#2$ & System $\#3$ \\
    \toprule
    
    Processor & Intel Core i9-7900X & 2 x Intel Xeon Gold 6154 & 2 x Intel Xeon Gold 6154 \\ 
    Base frequency & 3.30 GHz & 3.00 GHz & 3.00 GHz \\
    Physical cores & 10 & 36 & 36 \\ 
    Logical cores & 20 & 72 & 72 \\ 
    \midrule
    RAM & 128 GB DDR4 & 256 GB DDR4 & 256 GB DDR4 \\ 
    \midrule
    GPUs & 1 x NVidia GTX 1080Ti & 1 x NVidia RTX 2080Ti & 4 x NVidia RTX 2080Ti \\ 
    GPU memory & 11GB GDDR5X & 11GB GDDR6 & 11GB GDDR6 \\ 
    \midrule
    OS & Ubuntu 18.04 64-bit & Ubuntu 18.04 64-bit & Ubuntu 18.04 64-bit \\
    GPU drivers & NVidia 440.44 & NVidia 418.40 & NVidia 418.40 \\
    
    \bottomrule
\end{tabular}
\caption{Hardware setups used for profiling and performance measurements (Systems \#1 and \#2) and for large-scale experiments with self-play and PBT (System \#3).}
\label{tab:hw}
\end{table*}

\subsubsection{Environments}

We used three reinforcement learning domains for benchmarking: Atari, VizDoom, and DeepMind Lab. For Atari we simply chose Breakout with 4-framestack, although other environments exhibit almost identical throughput. The VizDoom scenario we selected is a simplified version of \textit{Battle} with a single discrete action head and the input space including only the pixel observations (no auxiliary game info). Most of the frameworks we tested do not support complex action and observation spaces, so this simplification allowed us to use the exact same version of the environment for all of the evaluated algorithms without major code modifications.

We chose \textit{rooms\_collect\_good\_objects\_train} from DMLab-30 as our benchmark environment for DeepMind Lab. This environment is also referred to as \textit{seekavoid\_arena\_01} in prior work \citeappendix{impala}. Just like the VizDoom scenario, this environment has pixel-based observations and a simple discrete action space.

In DeepMind Lab some environment states can be significantly harder to render, and therefore the simulation time depends on the behavior of the agent, e.g. as the agent learns to explore the environment the simulation can slow down or speed up as the distribution of visited states changes. To eliminate this potential source of variance in throughput we ignore the action distribution provided by the policy and sample actions randomly instead in our performance measurements for DMLab. This way we can measure only the throughput, disentangled from the learning performance. Note that using the random policy for acting does not change the amount of computation done by the algorithm. We collect and process the experience in the exact same way, only the actions sampled from the policy are replaced by random actions on the actors.

VizDoom environments are rendered with native resolution of $160 \times 120 \times 3$ which is downsampled to $128 \times 72 \times 3$. For DMLab the observation resolution is $96 \times 72 \times 3$. For VizDoom and DMLab we used 4-frameskip and no framestacking. Atari frames are rendered in $210 \times 160 \times 3$ and downsampled to $84 \times 84$ greyscale images. For Atari we used 4-frameskip and 4-framestack in all measurements, although higher overall throughput can be achieved without frame stacking.
Following \citeappendix{impala} and \citeappendix{seedrl} we report the throughput of all algorithms measured in \textit{environment frames} per second, i.e. a  number of simulated environment transitions, or, in our case, $4 \times$ the number of samples processed by the learner per second.

\subsubsection{Model architectures}

\begin{figure*}[ht]
\centering
\vspace{1mm}
\begin{subfigure}
  \centering
  {\small
  \def\svgwidth{0.47\textwidth}
  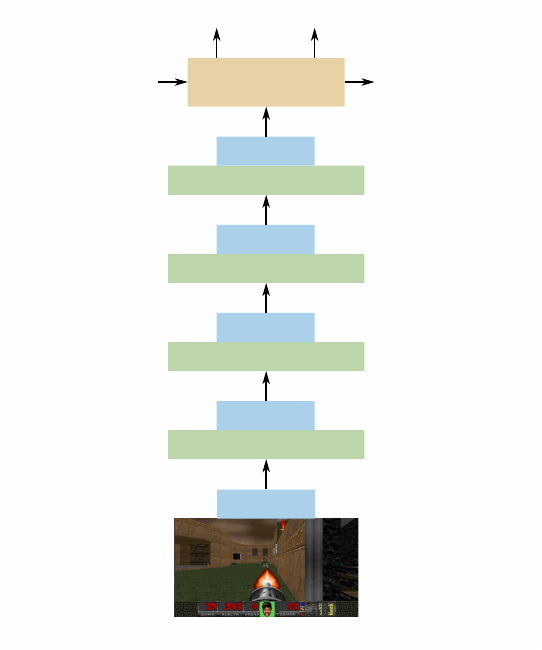
  }
\end{subfigure}
\begin{subfigure}
  \centering
  {\small
  \def\svgwidth{0.47\textwidth}
  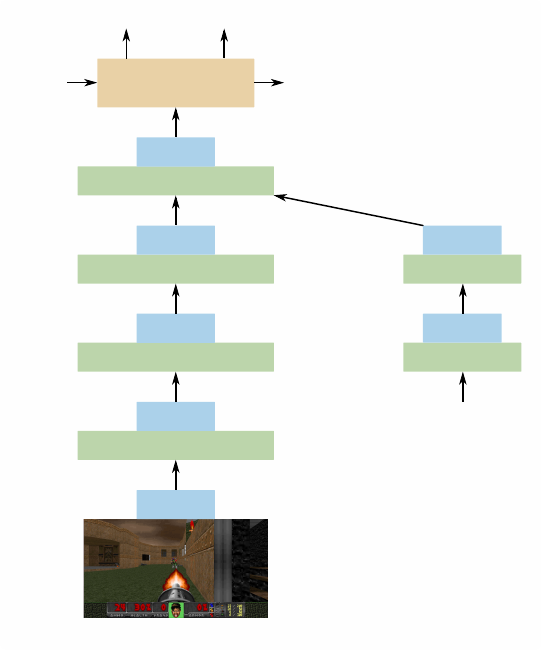
  }
\end{subfigure}
\caption{Neural network architectures used in VizDoom experiments. \textbf{Left:} \textit{simplified} architecture used for performance measurements and standard VizDoom environments. \textbf{Right:} \textit{full} architecture with additional low-dimensional game information input (health, armor, ammunition, etc.) and $L$ independent action heads. }
\label{fig:nn}
\end{figure*}

In all our performance benchmarks we used the same convolutional neural network to parameterize the actor and the critic, which is similar to model architectures used in prior work \citeappendix{a3c, impala}. In our implementation the 3-layer convolutional head is followed by a fully-connected layer, an LSTM core, and another pair of fully-connected layers to output the action distribution and the baseline. This architecture is referred to as \textit{simplified} (see Figure \ref{fig:nn}), in contrast to the \textit{full} architecture used in \textit{Battle}, \textit{Deathmatch}, and \textit{Duel} experiments, that contains additional observation and action spaces. We used the \textit{simplified} architecture to benchmark throughput in Atari, VizDoom, and DMLab.

Note that in our large-scale VizDoom experiments with the \textit{full} model we chose to use GRU RNN cells \citeappendix{gru_Kyunghyun} instead of LSTM \citeappendix{lstm_Hochreiter_97}. Empirically we find that GRU cells exhibit similar sample efficiency to LSTM cells and require slightly less computation.


\subsubsection{Benchmarking results}

We provide benchmarking results in the tabular form (see Table \ref{tab:throughput}). Data points are omitted for configurations that could not be initialized due to lack of resources, such as memory, simultaneously open file descriptors, or active parallel threads. Since Sample Factory allocates a very minimal amount of resources per environment instance, we were able to test configurations running as many as 3000 environments on a single machine for VizDoom and Atari, although increasing number of environments further provides diminishing returns.

Table \ref{tab:additional_benchmarks} shows performance figures for SampleFactory in some additional scenarios. As mentioned in the main paper, using GPU for rendering DeepMind Lab environments can improve performance, especially on systems with fewer CPU cores (e.g. System $\#1$).

Finally, we show performance figures for population-based training scenarios. Here we use 4 GPUs to accelerate learners and policy workers associated with up to 12 agents trained in parallel. Performance figures show that there is a very small penalty for increasing the population size, despite the fact that the amount of communication required grows significantly (e.g. rollout workers have to send observations to many different policy workers associated with different agents). The measurements in the table show the performance only for single-player environments. Multiplayer environments that involve actual network communication between individual game instances are significantly slower, up to 2-3 times, depending on the number of communicating instances. Significant performance gains are possible through replacing network communication between game instances with faster local mechanism, although this could require significant modifications to the VizDoom engine and lies beyond the scope of this project.

\begin{table*}[ht]
\vspace{1mm}
\setlength{\tabcolsep}{2.0pt}
\centering
\begin{threeparttable}
\centering
\scriptsize
\begin{tabular}{l p{0.5mm}<{\centering} cccccc p{0.5mm}<{\centering} cccccc p{0.5mm}<{\centering} cccccc}
  \multicolumn{22}{c}{\textbf{System 1 (10xCPU, 1xGPU)}} \\
  \specialrule{.1em}{.2em}{.5em} 

  && \multicolumn{6}{c}{Atari 84x84x4 } && \multicolumn{6}{c}{VizDoom 128x72 RGB} && \multicolumn{6}{c}{DmLab 96x72 RGB} \\

  \cmidrule(l{1mm}r{1mm}){3-8} \cmidrule(l{1mm}r{1mm}){10-15}
  \cmidrule(l{1mm}r{1mm}){17-22}
  
  \multicolumn{2}{r}{\# of envs sampled:}
  & 20 & 40 & 80 & 160 & 320 & 640 && 20 & 40 & 80 & 160 & 320 & 640 && 20 & 40 & 80 & 160 & 320 & 640 \\
  
  \midrule
  DeepMind IMPALA  && 6350 & 6470 & 6709 & 6880 & - & - && 6615 & 6776 & 7041 & 6669 & - & - && 6179 & 5943 & 6133 & 6448 & - & - \\
  SeedRL IMPALA  && 11347 & 15734 & 20715 & 24906 & 26149 & - && 11443 & 14537 & 19705 & 22059 & 22733 & - && 6747 & 10293 & 11262 & 11191 & 10604 & - \\
  RLLib IMPALA   && 10808 & 13596 & 17744 & 20236 & 21192 & 18232 && 10676 & 12556 & 12472 & 13444 & 11500 & 11868 && 7736 & 9224 & 9948 & 11644 & 11516 & - \\
  rlpyt PPO     && 13312 & 17764 & 21772 & 27240 & 31408 & 35272 && 16268 & 23688 & 26448 & 31660 & 38908 & 41940 && \textbf{9028} & 10852 & 11376 & 11560 & 12280 & 12400  \\
  \midrule
  SampleFactory APPO  && \textbf{17544} & \textbf{25307} & \textbf{35287} & \textbf{42113} & \textbf{46169} & \textbf{48016} && \textbf{16985} & \textbf{24809} & \textbf{37300} & \textbf{47913} & \textbf{55772} & \textbf{59525} && 8183 & \textbf{11792} & \textbf{12903} & \textbf{13040} & \textbf{13869} & \textbf{14746} \\

  \midrule \\
  \multicolumn{22}{c}{\textbf{System 2 (36xCPU, 1xGPU)}} \\
  \specialrule{.1em}{.2em}{.5em} 
  && 
  \multicolumn{6}{c}{Atari 84x84x4} &&
  \multicolumn{6}{c}{VizDoom 128x72 RGB} &&
  \multicolumn{6}{c}{DmLab 96x72 RGB} \\

  \cmidrule(l{1mm}r{1mm}){3-8} \cmidrule(l{1mm}r{1mm}){10-15} \cmidrule(l{1mm}r{1mm}){17-22}
  \multicolumn{2}{r}{\# of envs sampled:}
  & {72} & {144} & {288} & {576} & {1152} & {1728} && {72} & {144} & {288} & {576} & {1152} & {1728} && {72} & {144} & {288} & {576} & {1152} & {1728} \\
  \midrule
  DeepMind IMPALA  	&& 9661 & 8826 & 8602 & - & - & - && 10708 & 10043 & 9990 & - & - & - &&  8782 & 8622 & 8491 & - & - & - \\
  SeedRL IMPALA  && 25400 & 33425 & 39500 & 39726 & - & - && 23395 & 29591 & 34428 & - & - & - &&  22814 & 30354 & 32149 & 34773 & - & - \\
  RLLib IMPALA && 19148 & 20960 & 20440 & 19328 & 19360 & 22440 && 11471 & 11361  &  12144 & 11974 & 12098 & 12391 && 12536 &  13084 & 13932 & - & - & -
  \\
  rlpyt PPO && 24520 & 33544 & 39920 & 53112 & 63984 & 68880 && 37848 & 40040 & 57792 & 68644 & 71080 & 73544  && 22700 & 24140 & 29180 & 29424 & 32652 & 32948
  \\
  \midrule
  SampleFactory APPO  && \textbf{37061} & \textbf{59610} & \textbf{81247} & \textbf{95555} & \textbf{120355} & \textbf{135893}  && \textbf{38955} & \textbf{61223} & \textbf{79857} & \textbf{103658} & \textbf{131571} & \textbf{146551} && \textbf{26421}  &  \textbf{37088} & \textbf{41781} & \textbf{42149} & \textbf{41383} & \textbf{41784} 
  \\

  \bottomrule
\end{tabular}
\vspace{1mm}
\caption{Throughput of asynchronous RL methods measured in environment frames per second (samples per second $\times 4$).}
\label{tab:throughput}
\end{threeparttable}
\end{table*}

\begin{table*}[ht]
\vspace{1mm}
\centering
\setlength{\tabcolsep}{3mm}
\scriptsize
\begin{tabular}{l c c c r }
    \toprule
    \textbf{Hardware} & \textbf{Training scenario} & \textbf{Rollout workers} & \textbf{Total number of envs} & \textbf{Throughput, env. frames/sec} \\
    \toprule
    System $\#1$ & DMLab with GPU rendering & 20 & 160 & 17952 \\ 
    System $\#1$ & DMLab with GPU rendering & 20 & 320 & 18243 \\
    \midrule
    System $\#3$ & VizDoom \textit{Battle} PBT, \textit{full} model, 4 agents & 72 & 2304 & 153602 \\
    System $\#3$ & VizDoom \textit{Battle} PBT, \textit{full} model, 8 agents & 72 & 2304 & 154081 \\
    System $\#3$ & VizDoom \textit{Battle} PBT, \textit{full} model, 12 agents & 72 & 2304 & 146443 \\
    \bottomrule
\end{tabular}
\caption{Performance of Sample Factory in additional training scenarios.}
\label{tab:additional_benchmarks}
\end{table*}

\subsection{DMLab-30 experiment}

In this section we share our findings related to multi-task training on DMLab-30. Overall, we largely follow the same training procedure as the original IMPALA implementation, e.g. we used the exact same model based on ResNet backbone. We found however that seemingly subtle implementation details can significantly influence the learning performance.

One of the key choices when training on a multi-task benchmark like DMLab-30 with an asynchronous RL algorithm is whether to give different tasks the same amount of samples, or the same amount of compute. We follow \citeappendix{impala} and employ the second strategy. Just like the original implementation of IMPALA, we spawn an equal number of workers for every task (in our case 90 workers on a 36-core system, 3 workers per task) and let the OS schedule these processes. Note that this gives somewhat unfair advantage to tasks which render faster, since with the same amount of CPU time more samples can be generated for the faster environments. Sample Factory supports both training regimes, but we decided to go with the IMPALA strategy to ensure fair comparison of scores. Also, the throughput is higher in this mode. The authors argue that this implementation detail (distribution of compute resources across tasks) should be stated explicitly whenever different multi-task algorithms are compared.

\begin{figure}[hbt]
\centering
\includegraphics[width=\linewidth]{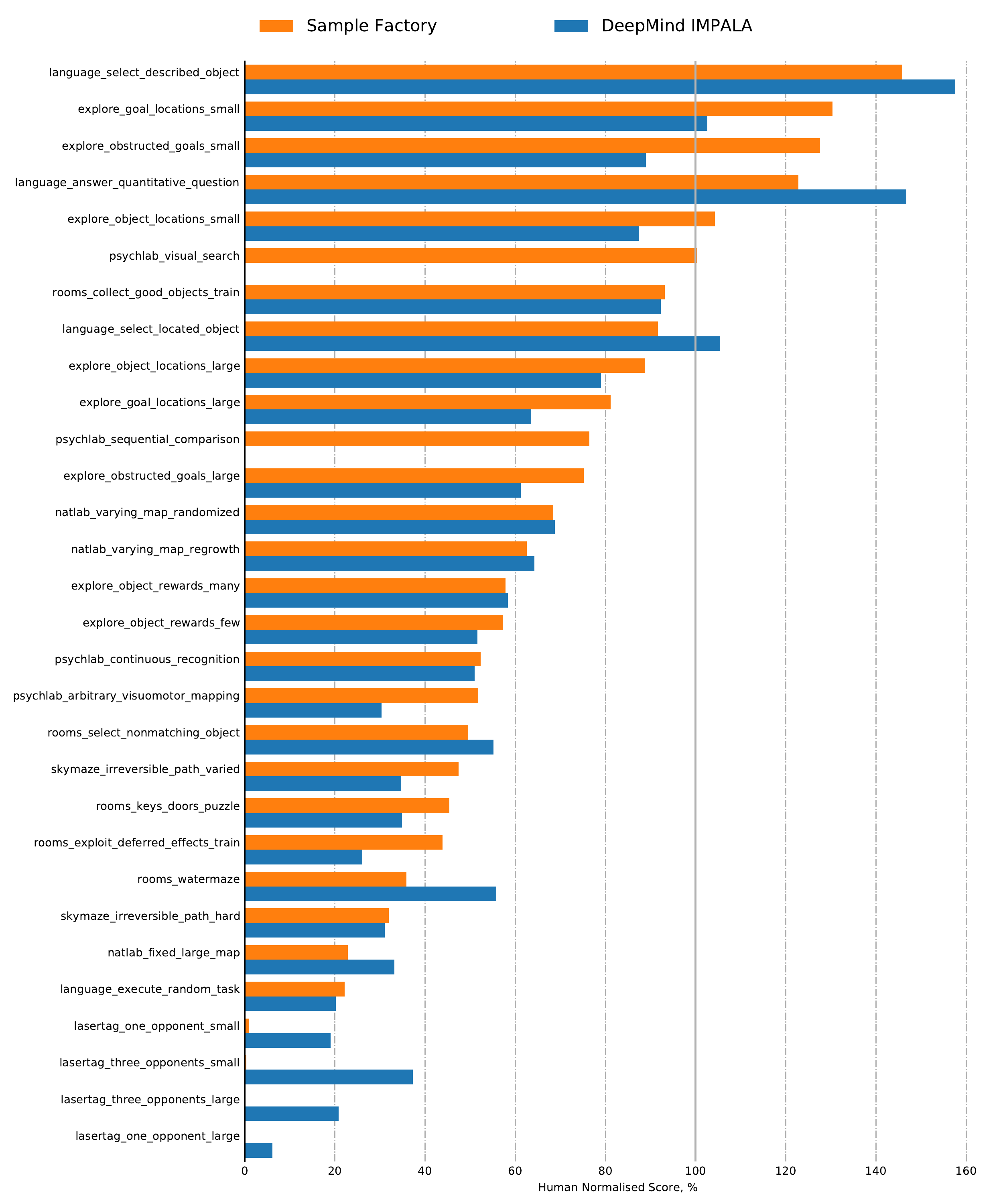}
\vspace{-5pt}
\caption{Final human-normalized training scores for individual DMLab-30 environments.}
\vspace{-5pt}
\label{fig:dmlab-30-individual-envs}
\end{figure}

The only significant difference compared to the original IMPALA setup is the chosen action space. We decided to use a slightly different discretization of the game inputs introduced in \citeappendix{popart_hessel_2019}, since it makes the action space closer to the one available to humans, e.g. it allows the agent to turn and move forward within the same frame. The increased number of actions, however, makes exploration harder, and we see a drop in performance in some of the levels where exploration is key. Figure \ref{fig:dmlab-30-individual-envs} shows the full breakdown of the agent's performance on individual tasks.

Finally, we noticed that one of the most significant factors affecting the throughput is the level generation at the episode boundary. To make the DMLab-30 benchmark more accessible we release a dataset of pre-generated levels, as well as the environment wrapper that makes it easy to use the dataset with any RL algorithm implementation. This wrapper builds on top of the already existing DMLab level cache. Without relying on the random seed provided by the environment, it will load the levels from the dataset until all of them are used in the training session, after which new levels will be generated and added to the cache. Follow the link below to find instructions on how to download the dataset and use it with Sample Factory: {\small\url{https://github.com/alex-petrenko/sample-factory#dmlab-level-cache}}.

\subsection{VizDoom experiments}

We used \textit{full} neural network architecture (as shown on Figure \ref{fig:nn}) to train our final VizDoom agents. All advanced VizDoom environments we used (\textit{Battle, Battle2, Deathmatch, Duel}) included an additional observation space with game information in numerical form. We only used information available to a human player through in-game UI. This includes: health and armor, current score, number of players in a match, selected weapon index, possession of different types of weapons, and amount of ammunition available for each weapon. We do not use previous rewards as a policy input, because the reward function can be based on hidden in-game information (e.g. damage dealt) and thus may give the agent an unfair advantage at test time.

Table \ref{tab:actions} describes the action space used in VizDoom experiments. We decompose the set of possible actions into seven independent action distributions, which allows the agent to combine multiple actions within the same frame, e.g. run forward, strafe, and attack at the same time. The action space for horizontal aim is technically continuous although in this work we discretize it with $1.25^{\circ}$ step, which empirically leads to faster learning.

\begin{table*}[ht]
\centering
\setlength{\tabcolsep}{3mm}
\small
\begin{tabular}{l @{\hspace{2em}} c @{\hspace{2em}} r }
    \toprule
    \textbf{Action head} & \textbf{Number of actions} & \textbf{Comment} \\
    \toprule
    Moving & 3 & no-action / forward / backward \\ 
    Strafing & 3 & no-action / left / right \\ 
    Attacking & 2 & no-action / attack \\ 
    Sprinting & 2 & no-action / sprint \\ 
    Object interaction & 2 & no-action / interact \\ 
    Weapon selection & 8 & no-action / select weapon slot 1..7 \\ 
    Horizontal aim & 21 & no-action / turning between $-12.5^{\circ}$ and $12.5^{\circ}$ in $1.25^{\circ}$ steps \\ 
    \midrule
    Total number of possible actions & 12096 & \\
    \bottomrule
\end{tabular}
\caption{Action space used in VizDoom multi-agent experiments.}
\label{tab:actions}
\end{table*}

Reward function for \textit{Battle} and \textit{Battle2} is based on the game score ($+1$ for killing a monster) plus a small additional reward for collecting health and ammo packs. In \textit{Deathmatch} and \textit{Duel} we extended the reward function to include penalties for dying, as well as additional rewards for picking up new types of weapons and dealing damage to opponents. Finally, we penalize the agent for switching the weapons too often, which accelerates the training in early stages.

The basic hyperparameters of all our experiments are presented in Table \ref{tab:hyperparams}. We deviate from these parameters only in \textit{Deathmatch} and \textit{Duel} experiments where we used action repeat (frameskip) of two consecutive frames instead of four. Consequently, we adjusted the discount factor to $0.995$ to account for this change. We observe that in these environments repeating actions fewer times led to better final performance of the agents.

In all hardware setups we used the number of rollout workers equal to the number of CPU cores. This allows us to use CPU affinity setting for processes to minimize the amount of context switching and accelerate sampling. The number of environments per core that enables the highest throughput lies between $2^4$ and $2^5$ for VizDoom. Note that for systems with large number of CPU cores a larger batch size might be required to reduce the policy lag. In all our experiments the policy lag was on average between 5 and 10 SGD steps, which results in stable training. Tensorboard summaries were used to monitor the policy lag during training.

\begin{table*}[ht]
\centering
\setlength{\tabcolsep}{3mm}
\small
\begin{tabular}{l | @{\hspace{3em}} r }
    \toprule
    Learning rate & $10^{-4}$ \\
    Action repeat (frameskip) & $2 / 4$ \\
    Framestack & No \\
    Discount $\gamma$ & $0.995 / 0.99$ \\
    Optimizer & Adam \citeappendix{adam} \\
    Optimizer settings & $\beta_1=0.9$, $\beta_2=0.999$, $\epsilon=10^{-6}$ \\
    Gradient norm clipping & $4.0$ \\
    \midrule
    Rollout length $T$ & 32 \\
    Batch size, samples & 2048 \\
    Number of training epochs & 1 \\
    \midrule
    V-trace parameters & $\Bar{\rho}=\Bar{c}=1$  \\
    PPO clipping range & $[1.1^{-1}, 1.1]$ \\
    \midrule
    Entropy coefficient & $0.003$ \\
    Critic loss coefficient & $0.5$ \\
    \bottomrule
\end{tabular}
\caption{Hyperparameters for VizDoom experiments.}
\label{tab:hyperparams}
\end{table*}

\subsubsection{Population-based training}

In our VizDoom population-based training experiments we used System $\#3$ to train a population of 8 agents in parallel. The full configuration of Sample Factory in this setup includes 72 rollout workers (one worker per logical core), 32 environment instances per rollout worker, 8 policy workers, and 8 learners (one for every policy involved). We deployed 2 learners and 2 policy workers on each available GPU with more GPU memory to spare.

Every $5M$ frames during training we randomly mutate hyperparameters and reward shaping weights of the bottom $70\%$ of the population. The mutation rate is $15\%$ for each hyperparameter. In our experiments we mutated learning rate, entropy loss coefficient, Adam $\beta_1$, and individual reward shaping coefficients by increasing or decreasing these parameters by a factor of $1.2$.
Additionally, every $5M$ frames we replace the policy weights for the worst $30\%$ of agents with weights of the policy randomly sampled from the best $30\%$. In \textit{Duel} experiment we introduce an additional threshold that prevents the weights exchange mechanism if policies are relatively close in performance (the difference in win rate is less than $0.35$), which helps to increase the diversity of the population.

\section{Additional performance considerations}

Seemingly small details can make a big difference in the performance of an asynchronous system. We found that tuning CPU core affinity and priority for various components of the system can give us a substantial performance gain. In Sample Factory we recommend setting the number of rollout workers to the number of logical CPU cores. In this case we can use processor affinity to run these worker processes on individual cores, preventing a lot of unnecessary context switching. We also found that in most configurations it helps to deprioritize rollout workers and let policy workers and learners to be scheduled as soon as there is any work available. This helps saturate the rollout workers with actions and increases the overall performance. Sample Factory comes with a default set of priorities/affinities that will work well for many training configurations.

In the highest throughput configurations batching of trajectories into minibatches and transferring them to the GPU can also become a bottleneck. Similar to \citeappendix{impala} and \citeappendix{seedrl} we implement this preprocessing step in the background thread on the learner, eliminating this particular performance issue.

\subsection{FIFO queues}

Sample Factory generally avoids explicit data transfer between system components, instead these components exchange addresses in shared memory buffers. Perhaps rather surprisingly, we found that at frame rates above $10^5$ FPS even communicating these addresses can be difficult. In fact, at this speed the standard Python's \textit{multiprocessing.Queue} tends to occupy a significant portion of CPU time. 

To solve this issue we implemented our own version of the IPC FIFO queue in C++, based on a circular buffer and POSIX mutexes. This custom implementation is a drop-in replacement for the standard \textit{multiprocessing.Queue} and it allows for 20-30 times faster message exchange in many producers - few consumers configuration, also achieving lower latency. The URL below contains installation instructions and detailed performance measurements:

{\small\url{https://github.com/alex-petrenko/faster-fifo}}

\bibliographyappendix{bibtex}
\bibliographystyleappendix{icml2020}

\end{document}